\pdfoutput=1
\documentclass[11pt]{article}
\usepackage{acl2015}
\usepackage{mathptmx}
\usepackage[scaled=.90]{helvet}
\usepackage{courier}
\usepackage{microtype}
\usepackage{latexsym}
\usepackage{array,arydshln}
\usepackage{multirow}
\usepackage{caption}
\usepackage{stfloats}
\usepackage{graphics}
\usepackage{pgfplots}
\usepackage{tikz}
\usepackage{amsmath}
\usepackage{array}
\title{Multi-domain Dialog State Tracking using Recurrent Neural Networks}

\author{Nikola Mrk\v{s}i\'c$^{\mathbf{1, 2}}$, ~ Diarmuid {\'O S\'eaghdha}$^{\mathbf{2}}$, ~ Blaise Thomson$^{\mathbf{2}}$, ~ Milica Ga\v{s}i\'c$^{\mathbf{1}}$   \\ \bf  ~ Pei-Hao Su$^{\mathbf{1}}$, ~ David Vandyke$^{\mathbf{1}}$, Tsung-Hsien Wen$^{\mathbf{1}}$ and Steve Young$^{\mathbf{1}}$  \\
 $^{\mathbf{1}}$  Department of Engineering, University of Cambridge, UK  \\
  $^{\mathbf{2}}$ VocalIQ Ltd. , Cambridge, UK \\
  {\small \tt \{nm480,mg436,phs26,djv27,thw28,sjy\}@cam.ac.uk ~ \small \tt\{diarmuid, blaise\}@vocaliq.com}}

\begin{document}
\maketitle
\begin{abstract} 
Dialog state tracking is a key component of many modern dialog systems, most of which are designed with a single, well-defined domain in mind. This paper shows that dialog data drawn from different dialog domains can be used to train a general belief tracking model which can operate across all of these domains, exhibiting superior performance to each of the domain-specific models. We propose a training procedure which uses out-of-domain data to initialise belief tracking models for entirely new domains. This procedure leads to improvements in belief tracking performance regardless of the amount of in-domain data available for training the model.  

\end{abstract}

\section{Introduction}

Spoken dialog systems allow users to interact with computer applications through a conversational interface. Modern dialog systems are typically designed with a well-defined domain in mind, e.g., restaurant search, travel reservations or shopping for a new laptop. The goal of building open-domain dialog systems capable of conversing about any topic remains far off. In this work, we move towards this goal by showing how to build \emph{dialog state tracking models} which can operate across entirely different domains. The state tracking component of a dialog system is responsible for interpreting the users' utterances and thus updating the system's \textit{belief state}: a probability distribution over all possible states of the dialog. This belief state is used by the system to decide what to do next.

Recurrent Neural Networks (RNNs) are well suited to dialog state tracking, as their ability to capture contextual information allows them to model and label complex dynamic sequences \cite{Graves:12}. In recent shared tasks, approaches based on these models have shown competitive performance \cite{Henderson:14b,Henderson:14d}. This approach is particularly well suited to our goal of building open-domain dialog systems, as it does not require handcrafted domain-specific resources for semantic interpretation. 

We propose a method for training {multi-domain RNN dialog state tracking models}. Our hierarchical training procedure first uses all the data available to train a {very general belief tracking model}. This model learns the most frequent and general dialog features present across the various domains. The general model is then specialised for each domain, learning domain-specific behaviour while retaining the cross-domain dialog patterns learned during the initial training stages. These models show robust performance across all the domains investigated, typically outperforming trackers trained on target-domain data alone. The procedure can also be used to initialise dialog systems for entirely new domains. In the evaluation, we show that such initialisation always improves performance, regardless of the amount of the in-domain training data available. We believe that this work is the first to address the question of multi-domain belief tracking.

\section{Related Work}

Traditional rule-based approaches to understanding in dialog systems (e.g.~ Goddeau et al.~\shortcite{Goddeau:EtAl:96}) have been superseded by data-driven systems that are more robust and can provide the probabilistic dialog state distributions that are needed by POMDP-based dialog managers. The recent Dialog State Tracking Challenge (DSTC) shared tasks \cite{Williams:13a,Henderson:14a,Henderson:14c} saw a variety of novel approaches, including robust sets of hand-crafted rules \cite{Wang:13}, conditional random fields \cite{Lee:13a,Lee:13b,Ren:13}, maximum entropy models \cite{Williams:13b} and web-style ranking \cite{Williams:14}.

Henderson et al.~\shortcite{Henderson:13,Henderson:14b,Henderson:14d} proposed a belief tracker based on recurrent neural networks. This approach maps directly from the ASR (automatic speech recognition) output to the belief state update, avoiding the use of complex semantic decoders while still attaining state-of-the-art performance. We adopt this RNN framework as the starting point for the work described here.

It is well-known in machine learning that a system trained on data from one domain may not perform as well when deployed in a different domain. Researchers have investigated methods for mitigating this problem, with NLP applications in parsing \cite{McClosky:EtAl:06,McClosky:EtAl:10}, sentiment analysis \cite{Blitzer:EtAl:07,Glorot:EtAl:11} and many other tasks. There has been a small amount of previous work on domain adaptation for dialog systems. Tur et al.~\shortcite{Tur:EtAl:07} and Margolis et al.~\shortcite{Margolis:EtAl:10} investigated domain adaptation for dialog act tagging. Walker et al.~\shortcite{Walker:EtAl:07} trained a sentence planner/generator that adapts to different individuals and domains. In the third DSTC shared task \cite{Henderson:14c}, participants deployed belief trackers trained on a restaurant domain in an expanded version of the same domain, with a richer output space but essentially the same topic. To the best of our knowledge, our work is the first attempt to build a belief tracker capable of operating across disjoint dialog domains.

\section{Dialog State Tracking using RNNs}
Belief tracking models capture users' goals given their utterances. Goals are represented as sets of constraints expressed by \emph{slot-value} mappings such as [food: \emph{chinese}] or [wifi: \emph{available}]. The set of slots $S$ and the set of values $V_s$ for each slot make up the \textit{ontology} for an application domain. 

Our starting point is the RNN framework for belief tracking that was introduced by Henderson et al.~\shortcite{Henderson:14b,Henderson:14d}. This is a single-hidden-layer recurrent neural network that outputs a distribution over all goal slot-value pairs for each user utterance in a dialog. It also maintains a \textit{memory} vector that stores internal information about the dialog context. The input for each user utterance consists of the ASR hypotheses, the last system action, the current memory vector and the previous belief state. Rather than using a spoken language understanding (SLU) decoder to convert this input into a meaning representation, the system uses the turn input to extract a large number of word $n$-gram features. These features capture some of the dialog dynamics but are not ideal for sharing information across different slots and domains.

\emph{Delexicalised $n$-gram features} overcome this problem by replacing all references to slot names and values with generic symbols. Lexical $n$-grams such as [{want cheap price}] and [{want Chinese food}] map to the same delexicalised feature, represented by [{want} \emph{tagged-slot-value} \emph{tagged-slot-name}]. Such features facilitate transfer learning between slots and allow the system to operate on unseen values or entirely new slots. As an example, [{want available internet}] would be delexicalised to [{want} \emph{tagged-slot-value} \emph{tagged-slot-name}] as well, a useful feature even if there is no training data available for the \emph{internet} slot. {The delexicalised model learns the belief state update corresponding to this feature from its occurrences across the other slots and domains. Subsequently, it can apply the learned behaviour to slots in entirely new domains}.

The system maintains a separate belief state for each slot $s$, represented by the distribution $\mathbf{p}_{s}$ over all possible slot values $v \in V_s$. The model input at turn $t$, $\mathbf{x}^{t}$, consists of the previous belief state $\mathbf{p}^{t-1}_{s}$, the previous memory state $\mathbf{m}^{t-1}$, as well as the vectors $\mathbf{ f}_l$ and $\mathbf{f}_{d}$ of lexical and delexicalised features extracted from the turn input\footnote{Henderson et al.'s work distinguished between three types of features: the delexicalised feature sets $\mathbf{f_{s}}$ and $\mathbf{f_{v}}$ are subsumed by our delexicalised feature vector $\mathbf{f}_{d}$, and the turn input $\mathbf{f}$ corresponds to our lexical feature vector $\mathbf{f}_{l}$.}. The belief state of each slot $s$ is updated for each of its slot values $v \in V_{s}$. The RNN memory layer is updated as well. The updates are as follows\footnote{The original RNN architecture had a second component which learned mappings from lexical $n$-grams to specific slot values. In order to move towards domain-independence, we do not use this part of the network.}:
\begin{align*} \mathbf{ x}^{t}_{v} ~ & = ~    \mathbf{f}_{l}^{t}  ~ \oplus ~ \mathbf{f}_{d}^{t}  ~ \oplus ~ { \mathbf{ m}^{t-1} } ~ \oplus ~ p^{t-1}_{v} ~ \oplus ~ p^{t-1}_{\emptyset}  \\
    g^{t}_{v}  ~ & = ~     \mathbf{w}_{1}^{s} \cdot \sigma \left( \mathbf{W}_{0}^{s} \mathbf{x}_{v}^{t}   + b_{0}^{s}   \right) + b_{1}^{s} \\
	p^{t}_{v}~ & = ~     \frac{\exp( g^{t}_{v} ) }{\exp(g_{\emptyset}^{t}) + \sum_{v' \in V} \exp(g^{t}_{v'}) } \\    
    \mathbf{ m} ^{t}~ & = ~     \sigma \left( \mathbf{ W}_{m_0}^{s} \mathbf{x}_{t} + \mathbf{W}_{m_1}^{s} \mathbf{m}^{t-1} \right) \end{align*} 
\noindent where $\oplus$ denotes vector concatenation and $p_{\emptyset}^{t}$ is the probability that the user has expressed no constraint up to turn $t$. Matrices $\mathbf{W}_{0}^{s}$, $\mathbf{ W}_{m_0}^{s}$, $\mathbf{W}_{m_1}^{s}$ and the vector $\mathbf{w}_{1}^{s}$ are the RNN weights, and $b_{0}$ and $b_{1}$ are the hidden and output layer RNN bias terms. 

For training, the model is unrolled across turns and trained using backpropagation through time and stochastic gradient descent \cite{Graves:12}. 

\section{Hierarchical Model Training}

Delexicalised features allow transfer learning between slots. We extend this approach to achieve transfer learning between domains: a model trained to talk about hotels should have some success talking about restaurants, or even laptops. If we can incorporate features learned from different domains into a single model, this model should be able to track belief state across all of these domains.

The training procedure starts by performing \emph{shared initialisation}: the RNN parameters of all the slots are tied and all the slot value occurrences are replaced with a single generic tag. These slot-agnostic delexicalised dialogs are then used to train the parameters of the \emph{shared RNN model}. 

Extending shared initialisation to training across multiple domains is straightforward. We first delexicalise all slot value occurrences for all slots across the different domains in the training data. This combined (delexicalised) dataset is then used to train the multi-domain shared model.

The shared RNN model is trained with the purpose of extracting a very rich set of lexical and delexicalised features which capture general dialog dynamics. While the features are general, the RNN parameters are not, since not all of the features are equally relevant for different slots. For example, [eat \emph{tagged-slot-value} food] and [near \emph{tagged-slot-value}] are clearly features related to \emph{food} and \emph{area} slots respectively. To ensure that the model learns the relative importance of different features for each of the slots, we train {slot specific models} for each slot across all the available domains. To train these \emph{slot-specialised} models, the shared RNN's parameters are replicated for each slot and specialised further by performing additional runs of stochastic gradient descent using only the slot-specific (delexicalised) training data. 

\section{Dialog domains considered}

We use the experimental setup of the Dialog State Tracking Challenges. The key metric used to measure the success of belief tracking is \emph{goal accuracy}, which represents the ability of the system to correctly infer users' constraints. We report the \emph{joint goal accuracy}, which represents the marginal test accuracy across all slots in the domain. 

We evaluate on data from six domains, varying across topic and geographical location (Table 1). The Cambridge Restaurants data is the data from DSTC 2. The San Francisco Restaurants and Hotels data was collected during the Parlance project \cite{Gasic:14}. The Tourist Information domain is the DSTC 3 dataset: it contains dialogs about hotels, restaurants, pubs and coffee shops. 

The Michigan Restaurants and Laptops datasets are collections of dialogs sourced using Amazon Mechanical Turk. The Laptops domain contains conversations with users instructed to find laptops with certain characteristics. This domain is substantially different from the other ones, making it particularly useful for assessing the quality of the multi-domain models trained. 

We introduce three \emph{combined} datasets used to train increasingly general belief tracking models:
\begin{enumerate}
\item \emph{All Restaurants} model: trained using the combined data of all three restaurant domains;
\item R+T+H model: trained on all dialogs related to restaurants, hotels, pubs and coffee shops;
\item R+T+H+L model: the most general model, trained using all the available dialog data. 
\end{enumerate}

\setcounter{table}{0}

\begin{table} [!tp]
\resizebox{1\columnwidth}{!}{%
\begin{tabular}{ccccc}
\bf Dataset / Model & \bf Domain & \bf Train & \bf Test & \bf Slots \\ \hline
\bf Cambridge Rest. & { Restaurants } &  2118 & 1117 & 4 \\ 
\bf SF Restaurants &  { Restaurants} & 1608 & 176 & 7 \\ 

\bf Michigan Rest. &  { Restaurants} & 845 & 146 & 12 \\  \hdashline
\bf All Restaurants &  { Restaurants} & 4398 & \bf - & 23 \\ \hline 
\bf Tourist Info. &  { Tourist Info}  & 2039 & 225 & 9 \\ 
\bf SF Hotels &  { Hotels Info} & 1086 & 120 & 7 \\ \hdashline
\bf R+T+H Model & {Mixed} & 7523 & \bf - & 39 \\ \hline
\bf Laptops & {Laptops} & 900 & 100 & 6 \\ \hline
\bf R+T+H+L Model & {Mixed} & 8423 & \bf - & 45 
\end{tabular}%
}%
\vspace{-2mm} \caption{\label{tab:datasets} datasets used in our experiments \vspace{-2mm}}

\end{table}

\section{Results}

As part of the evaluation, we use the three combinations of our dialog domains to build increasingly \emph{general} belief tracking models. The domain-specific models trained using only data from each of the six dialog domains provide the baseline performance for the three {general models.}

\setcounter{table}{1}
\begin{table*} [htp!]
\begin{center}
\resizebox{2.05\columnwidth}{!}{%
\begin{tabular}{c|cccccc|c}
\bf { Model / Domain } & \small{\bf Cam Rest} &  \small{\bf SF Rest} & \small{\bf Mich Rest} & \small{\bf Tourist}   & \small{\bf SF Hotels} & \small{\bf Laptops} & \small{\bf Geo. Mean }  \\ \hline

\bf  { Cambridge Restaurants } &{ 75.0 } & { 26.2 } & { 33.1 } & { 48.7 } & { 5.5 } & { 54.1 } & { 31.3 } \\ 
\bf  { San Francisco Restaurants } &{ 66.8 } & { \bf 51.6 } & { 31.5 } & { 38.2 } & { 17.5 } & { 47.4 } & { 38.8 } \\ 
\bf  { Michigan Restaurants } &{ 57.9 } & { 22.3 } & { 64.2 } & { 32.6 } & { 10.2 } & { 45.4 } & { 32.8 } \\ \hdashline
\bf  { All Restaurants } &{ 75.5 } & { 49.6 } & { 67.4 } & { 48.2 } & { 19.8 } & { 53.7 } & { 48.5 } \\ \hline
\bf  { Tourist Information } &{ 71.7 } & { 27.1 } & { 31.5 } & { 62.9 } & { 10.1 } & { 55.7 } & { 36.0 } \\ 
\bf  { San Francisco Hotels } &{ 26.2 } & { 28.7 } & { 27.1 } & { 27.9 } & { 57.1 } & { 25.3 } & { 30.6 } \\  \hdashline
\bf  { Rest $\cup$ Tourist $\cup$ Hotels (R+T+H) } &{ \bf 76.8 } & { 51.2 } & { \bf 68.7 } & { \bf 65.0 } & { \bf 58.8 } & { 48.1 } & { 60.7 } \\  \hline
\bf  { Laptops } &{ 66.9 } & { 26.1 } & { 32.0 } & { 46.2 } & { 4.6 } & { 74.7 } & { 31.0 } \\ \hdashline
\bf  { All Domains (R+T+H+L) } &{ \bf 76.8 } & { 50.8 } & { 64.4 } & { 63.6 } & { 57.8 } & { \bf 76.7 } & { \bf 64.3 } \\

\end{tabular}%
} \vspace{-0mm}
\caption{\label{font-table} Goal accuracy of shared models trained using different dialog domains (ensembles of 12 models)\vspace{-4mm}} 
\end{center}
\end{table*}

\setcounter{table}{2}

\begin{table*} [!hbp]
\begin{center}

\resizebox{2.05\columnwidth}{!}{%
\begin{tabular}{c|cc|cc|cc}

\multirow{2}{*}{ \bf \large Model} & \multicolumn{2}{c|}{\bf Cambridge Restaurants}& \multicolumn{2}{c|}{\bf SF Restaurants} & \multicolumn{2}{c}{\bf Michigan Restaurants} \\ 

 & \small{ Shared Model} & \small{ Slot-specialised} & \small{Shared Model} & \small{Slot-specialised}  &  \small{ Shared Model} & \small {Slot-specialised}  \\ 

\hline

\bf ~~Domain Specific~~ & { 75.0  } & {  75.4  }   & 51.6  & \bf 56.5   & 64.2  & 65.6  \\    
\bf  All Restaurants &   75.5 & 77.3 &  49.6 & 53.6 & 67.4 & 65.9   \\ 
\bf   R+T+H  &  {   76.8 } & { \bf 77.4  }  &  51.2 &  54.6  & \bf 68.7 & 65.8 \\ 
\bf R+T+H+L  &  {  76.8  } & {77.0}  & 50.8 & 54.1 & 64.4  &  66.9 \\ \hline

\multirow{2}{*}{ \bf \large } & \multicolumn{2}{c|}{\bf Tourist Information}& \multicolumn{2}{c|}{\bf SF Hotels} & \multicolumn{2}{c}{\bf Laptops} \\ 

 & \small{ Shared Model} & \small{ Slot-specialised} & \small{Shared Model} & \small{ Slot-specialised} &  \small{ Shared Model} & \small{  Slot-specialised}  \\ 

\hline

\bf   ~~Domain Specific~~  & { 62.9 } &  {  65.1 }    &  57.1  &   57.4   &  74.7 &  78.4   \\ 
\bf   R+T+H  &  {  65.0 } &  {  \bf 67.1 }  &    58.8 &  60.7 &    -  &   - \\ 
\bf   R+T+H+L &  { 63.6 } &  {65.5}  &  {57.8} &  \bf 61.6 &    76.7 &   \bf 78.9  \\
\end{tabular}%
}

\vspace{-1mm} \caption{Impact of slot specialisation on performance across the six domains (ensembles of 12 models)  \label{tab:specialisation}}
\end{center}

\end{table*}

\subsection{Training General Models}

Training the shared RNN models is the first step of the training procedure. Table 2 shows the performance of shared models trained using dialogs from the six individual and the three combined domains. The joint accuracies are not comparable between the domains as each of them contains a different number of slots. The \emph{geometric mean} of the six accuracies is calculated to determine how well these models operate across different dialog domains.  

The parameters of the three multi-domain models are not slot or even domain specific. Nonetheless, all of them improve over the domain-specific model for all but one of their constituent domains. The {R+T+H} model outperforms the R+T+H+L model across four domains, showing that the use of laptops-related dialogs decreases performance slightly across other more closely related domains. However, the latter model is much better at balancing its performance across all six domains, achieving the highest geometric mean and still improving over all but one of the domain-specific models.

\subsection{Slot-specialising the General Models}

\emph{Slot specialising} the shared model allows the training procedure to learn the relative importance of different delexicalised features for each slot in a given domain. Table 3 shows the effect of slot-specialising shared models across the six dialog domains. Moving down in these tables corresponds to adding more out-of-domain training data and moving right corresponds to slot-specialising the shared model for each slot in the current domain. 

Slot-specialisation improved performance in the vast majority of the experiments. All three slot-specialised general models outperformed the RNN model's performance reported in DSTC 2.

\subsection{Out of Domain Initialisation}

The hierarchical training procedure can exploit the available out-of-domain dialogs to initialise improved shared models for new dialog domains.

In our experiments, we choose one of the domains to act as the \emph{new} domain, and we use a subset of the remaining ones as \emph{out-of-domain} data. The number of in-domain dialogs available for training is increased at each stage of the experiment and used to train and compare the performance of two slot-specialised models. These models slot-specialise from two different shared models. One is trained using in-domain data only, and the other is trained on all the out-of-domain data as well. 

The two experiments vary in the degree of similarity between the in-domain and out-of-domain dialogs. In the first experiment, Michigan Restaurants act as the new domain and the remaining R+T+H dialogs are used as out-of-domain data. In the second experiment, Laptops dialogs are the in-domain data and the remaining dialog domains are used to initialise the more general shared model. 

\begin{figure*} [!ht]
\centering
\begin{minipage}{1.03\columnwidth}
\centering
\begin{tikzpicture}
\begin{axis}[
x tick label style={ /pgf/number format/1000 sep=},
legend pos=south east,
legend cell align=left,
scaled y ticks=true,
xmin=-50,ymin=49,
legend entries={\small{In-domain Initialisation}, \small{Out-of-domain Initialisation}}
]
\addplot[blue,dashed,line width=0.3mm] table {ensemble_4_michigan.dat};
\addplot[red,smooth,line width=0.3mm] table {ensemble_4_slotspec_mr.dat};

\end{axis}

\end{tikzpicture}\end{minipage}\hfill
\begin{minipage}{1.03\columnwidth}
\centering
\begin{tikzpicture}
\begin{axis}[
x tick label style={ /pgf/number format/1000 sep=},
legend pos=south east,
legend cell align=left,
scaled y ticks=true,
xmin=-50,ymin=24,ymax=85,
legend entries={\small{In-domain Initialisation}, \small{Out-of-domain Initialisation}}
]
\addplot[blue,dashed,line width=0.3mm] table {ensemble_4_laptops.dat};
\addplot[red,smooth,line width=0.3mm] table {ensemble_4_slotspec_laptops.dat};

\end{axis}
\end{tikzpicture}
\end{minipage}
\vspace{2mm} \caption{Joint goal accuracy on Michigan Restaurants (left) and the Laptops domain (right) as a function of the number of in-domain training dialogs available to the training procedure (ensembles of four models)\label{fig:michigan} \vspace{-4mm} }
\end{figure*}
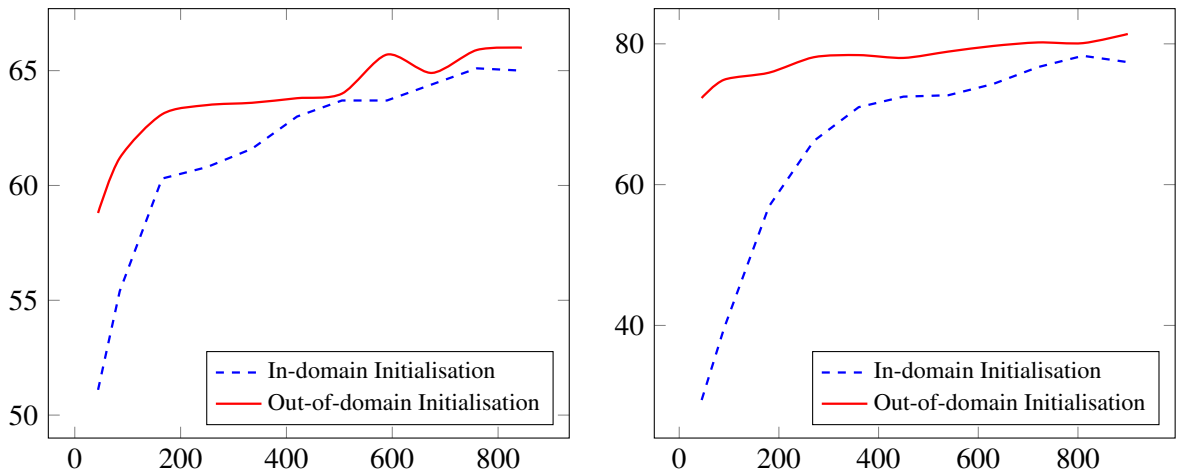

Figure 1 shows how the performance of the two differently initialised models improves as additional in-domain dialogs are introduced. In both experiments, the use of out-of-domain data helps to initialise the model to a much better starting point when the in-domain training data set is small. The out-of-domain initialisation consistently improves performance: the joint goal accuracy is improved even when the entire in-domain dataset becomes available to the training procedure.

These results are not surprising in the case of the system trained to talk about Michigan Restaurants. Dialog systems trained to help users find restaurants or hotels should have no trouble finding restaurants in alternative geographies. In line with these expectations, the use of a shared model initialised using R+T+H dialogs results in a model with strong starting performance. As additional restaurants dialogs are revealed to the training procedure, this model shows relatively minor performance gains over the domain-specific one. 

The results of the Laptops experiment are even more compelling, as the difference in performance between the differently initialised models becomes larger and more consistent. There are two factors at play here: exposing the training procedure to substantially different out-of-domain dialogs allows it to learn delexicalised features not present in the in-domain training data. These features are applicable to the Laptops domain, as evidenced by the very strong starting performance. As additional in-domain dialogs are introduced, the delexicalised features not present in the out-of-domain data are learned as well, leading to consistent improvements in belief tracking performance. 

In the context of these results, it is clear that the out-of-domain training data has the potential to be even more beneficial to tracking performance than data from relatively similar domains. This is especially the case when the available in-domain training datasets are too small to allow the procedure to learn appropriate delexicalised features.

\section{Conclusion}
		
We have shown that it is possible to train general belief tracking models capable of talking about many different topics at once. The most general model exhibits robust performance across {all domains}, outperforming most domain-specific models. This shows that training using diverse dialog domains allows the model to better capture general dialog dynamics applicable to different domains at once. 

The proposed hierarchical training procedure can also be used to adapt the general model to new dialog domains, with very small in-domain data sets required for adaptation. This procedure improves tracking performance even when substantial amounts of in-domain data become available. 

\subsection{Further Work}
The suggested domain adaptation procedure requires a small collection of annotated in-domain dialogs to adapt the general model to a new domain. In our future work, we intend to focus on initialising good belief tracking models when no annotated dialogs are available for the new dialog domain.

\bibliographystyle{acl}

\bibliography{references_all}{}

\end{document}